\documentclass[final]{cvpr}

\usepackage{times}
\usepackage{epsfig}
\usepackage{graphicx}
\usepackage{amsmath}
\usepackage{amssymb}
\usepackage{booktabs}
\usepackage{multirow}
\usepackage{comment}
\usepackage{subfigure}

\usepackage[pagebackref=true,breaklinks=true,colorlinks,bookmarks=false]{hyperref}
\usepackage{array}

    \makeatletter
    \newcommand{\thickhline}{%
        \noalign {\ifnum 0=`}\fi \hrule height 1pt
        \futurelet \reserved@a \@xhline
}
    \newcolumntype{"}{@{\vrule width 1pt}}
    \makeatother

\newcommand{\tabincell}[2]{\begin{tabular}{@{}#1@{}}#2\end{tabular}}

\def\cvprPaperID{2823} 
\def\confYear{CVPR 2021}

\begin{document}

\title{UnrealPerson: An Adaptive Pipeline towards Costless Person Re-identification}

\author{
Tianyu Zhang,\textsuperscript{\rm 1} Lingxi Xie,\textsuperscript{\rm 4} Longhui Wei,\textsuperscript{\rm 5} Zijie Zhuang,\textsuperscript{\rm 4} Yongfei Zhang,\textsuperscript{\rm 1,2,3} Bo Li,\textsuperscript{\rm 1,2,3} Qi Tian \textsuperscript{\rm 6}\\
\small
\textsuperscript{\rm 1}Beijing Key Laboratory of Digital Media, School of Computer Science and Engineering, Beihang University, \\
\small
\textsuperscript{\rm 2}State Key Laboratory of Virtual Reality Technology and Systems, Beihang University,\\
\small
\textsuperscript{\rm 3}Pengcheng Laboratory,
\textsuperscript{\rm 4}Tsinghua University,
\small
\textsuperscript{\rm 5}University of Science and Technology of China,
\textsuperscript{\rm 6}Xidian University\\
\small
zhangtianyu@buaa.edu.cn, 198808xc@gmail.com, longhuiwei@pku.edu.cn, \\
\small
jayzhuang42@gmail.com, yfzhang@buaa.edu.cn, boli@buaa.edu.cn, q-tian@hotmail.com
}

\maketitle

\noindent\footnotetext{This work was partially supported by the National Key R\&D Program of China (Grant No. 2020AAA0130200), the National Natural Science Foundation of China (No. 61772054,  62072022), the NSFC Key Project (No. 61632001) and the Fundamental Research Funds for the Central Universities.}

\begin{abstract}

%

The main difficulty of person re-identification (ReID) lies in collecting annotated data and transferring the model across different domains. This paper presents \textbf{UnrealPerson}, a novel pipeline that makes full use of unreal image data to decrease the costs in both the training and deployment stages. Its fundamental part is a system that can generate synthesized images of high-quality and from controllable distributions. Instance-level annotation goes with the synthesized data and is almost free. We point out some details in image synthesis that largely impact the data quality.
With 3,000 IDs and 120,000 instances, our method achieves a 38.5\% rank-1 accuracy when being directly transferred to MSMT17.
It almost doubles the former record using synthesized data and even surpasses previous direct transfer records using real data.
This offers a good basis for unsupervised domain adaption, where our pre-trained model is easily plugged into the state-of-the-art algorithms towards higher accuracy. In addition, the data distribution can be flexibly adjusted to fit some corner ReID scenarios, which widens the application of our pipeline.
We will publish our data synthesis toolkit and synthesized data in \url{https://github.com/FlyHighest/UnrealPerson}.

\end{abstract}

\section{Introduction}

Person re-identification aims to retrieve the same pedestrian (\textit{i.e.}, an identity) from the images captured by a camera network. As a fundamental task of intelligent surveillance, ReID has attracted increasing attention in the computer vision community. Recently, with the emergence of large-scale ReID datasets~\cite{MARKET,DUKEMTMC,PTGAN}, effective algorithms have been proposed and achieved satisfying performance in these benchmarks. However, there is still a significant gap in deploying the ReID algorithms to real-world scenarios, arguably because
(i) the trained models are often vulnerable to domain changes, yet (ii) annotating identities in new scenarios requires exhausting human labors.
We attribute such an application gap to the fact that the current pipeline is hindered by the data limitation and thus not optimized for generalizing across different scenarios.


To alleviate this problem and pave a new path for the community, we propose \textbf{UnrealPerson}, a new pipeline that makes full use of unreal (synthesized) image data towards a powerful ReID algorithm that easily and costlessly deploys to a wide range of scenarios. The key observation is that the synthesized pedestrian data sampled from a virtual world comes naturally with free and perfect annotations. From the perspective of the generalization ability, the synthesized data enjoys two-fold benefits.
\textbf{First}, compared to the manually collected data from restricted real scenarios, the synthesized data from infinite virtual scenes is more diverse, making it less prone to domain-specific patterns.
\textbf{Second}, the parameters of data synthesis can be freely adjusted to fit the domains in which collecting real data is difficult (\textit{e.g.}, the low-illumination scenario).
To fully utilize these characteristics, our entire pipeline consists of \textbf{\textit{pre-training the model using abundant synthesized data and then fine-tuning it with off-the-shelf domain adaptation algorithms}}.
This pipeline has broad applications since it fits both the fully-supervised and unsupervised setting and transfers well to a few corner scenarios.

The quality and richness of our synthesized pedestrian data is the cornerstone of our UnrealPerson pipeline. 
To synthesize abundant and authentic samples, we first create a set of scenarios (\textit{e.g.}, street, plaza, \textit{etc.}) in the virtual 3D world with changeable environmental parameters (\textit{e.g.}, illumination, lighting, \textit{etc.}). Then, we place an arbitrary number of pedestrians with configurable appearance (\textit{e.g.}, gender, height, clothing, \textit{etc.}) into the scenarios, and they move according to the pre-defined paths. Finally, the images are captured by the virtual cameras in the scenes.
With this data synthesis system, UnrealPerson is flexible to assemble suitable training data and learning approaches to achieve the best performance for different ReID tasks.
We verify its effectiveness through two groups of experiments.
\textbf{First}, we verify the effectiveness of the synthesized data on improving the generalization ability.
We train the ReID model only on the synthesized data and test it directly on conventional ReID benchmarks.
Quantitatively, our vanilla baseline~\cite{CBN} achieves a rank-1 accuracy of $38.5\%$ on the MSMT17 dataset, which almost doubles the previous synthesis-based record: $20.0\%$ by RandPerson~\cite{RandPerson}.
\textbf{Second}, we adapt our pipeline for specialized ReID scenarios, \textit{e.g.}, all pedestrians are in black, or the illumination is very low.
In these tough scenarios, UnrealPerson achieves competitive performance, even surpassing the models that are pre-trained in manually labeled datasets with real-world images.
Our major contribution is summarized as follows.
\begin{itemize}
\setlength{\itemsep}{0.01cm}
\item We propose a novel pipeline that largely reduces the deployment costs of ReID. For the first time, the model pre-trained purely on synthesized data outperforms that pre-trained on real, annotated data.
\setlength{\itemsep}{0.01cm}
\item We verify the usefulness of our pipeline in a wide range of downstream tasks, including direct transfer, supervised domain adaptation, and unsupervised domain adaptation settings.
\setlength{\itemsep}{0.01cm}
\item We provide a detailed analysis of the factors in data synthesis, which offers practical guidelines for reusing our toolkit for future research.

\end{itemize}

\section{Related Work}

\subsection{ReID: Full Supervision and Direct Transfer}
A typical ReID framework for supervised learning requires annotating identities, and hence deep networks can be optimized by learning to separate different persons.
The methods can be roughly classified into two categories,~\emph{i.e.}, improving feature extraction and designing better objective functions.
To make the feature more robust to pose variations, part-based methods~\cite{Sun2018_beyond,PSE,GLAD,PART_ALIGNED_REID,zhao2017spindle} are proposed. These methods extract local features from body parts to complement the global features.
Some methods target to enhance person-related feature extraction by eliminating the background interference via semantic parsing~\cite{Elimi} or attention mechanism~\cite{MASKREID,han}.
Also, some works~\cite{DeepReID,MGN,autoreid} contribute to more effective network architecture for person feature extraction.
Moreover, powerful objective functions are introduced to ReID, including triplet loss~\cite{InDefense,point2set}, contrastive loss~\cite{gated}, center loss~\cite{bagoftricks,centerloss}, circle loss~\cite{circleloss}, and so on.
These methods achieve good performance on the same domain evaluation but report unsatisfactory results when directly transferred to other domains.
Some methods aim to overcome over-fitting on the training domain and make person ReID models more generalizable. For example, Liao~\emph{et al.}~\cite{QAConv} conduct pairwise matching to find explainable local similar regions. Song~\emph{et al.}~\cite{DIMN} follow meta-learning pipelines and extract domain-invariant features via sampled sub-domains.
Zhuang~\emph{et al.}~\cite{CBN} propose to align the feature distribution of all cameras, which also benefits direct transfer evaluation.
However, the domain gap limits the performance of these methods.

\subsection{Domain Adaptation for Person ReID}
Domain adaptation on the target domain usually boosts the performance significantly by shrinking the huge domain gap.
Fine-tuning with annotated data can be regarded as a basic supervised domain adaptation method, which is widely used in many computer vision tasks~\cite{He_2019_ICCV}.
Further, Xiao~\emph{et al.}~\cite{DGD} propose to fine-tune the pre-trained model with domain-guided dropout to filter out useless neurons.
On the other hand, unsupervised domain adaptation (UDA) attracts more attention because it requires cheaper unlabeled data of the target domain.
The means of UDA include data augmentation~\cite{PTGAN,SPGAN,ECN,cameramixup}, distribution alignment~\cite{D-MMD,GDS}, predicting pseudo labels~\cite{JVTC,SpCL,DCML,BottomUp,softlabel,MMT}, spatial-temporal consistency mining~\cite{tfusion,JVTC}, model ensemble~\cite{MEB-Net,NRMT}, and so on.
The final performance of UDA also relies on the transferability of pre-training data.
For example, pseudo label based methods rely on the pre-trained model to provide the initial labels, and the accuracy of pseudo labels directly influences the model convergence and the final performance.
Therefore, the quality of pre-training data is the basis for domain adaptation.

\begin{figure*}[ht!]
    \centering
    \includegraphics[width=\textwidth]{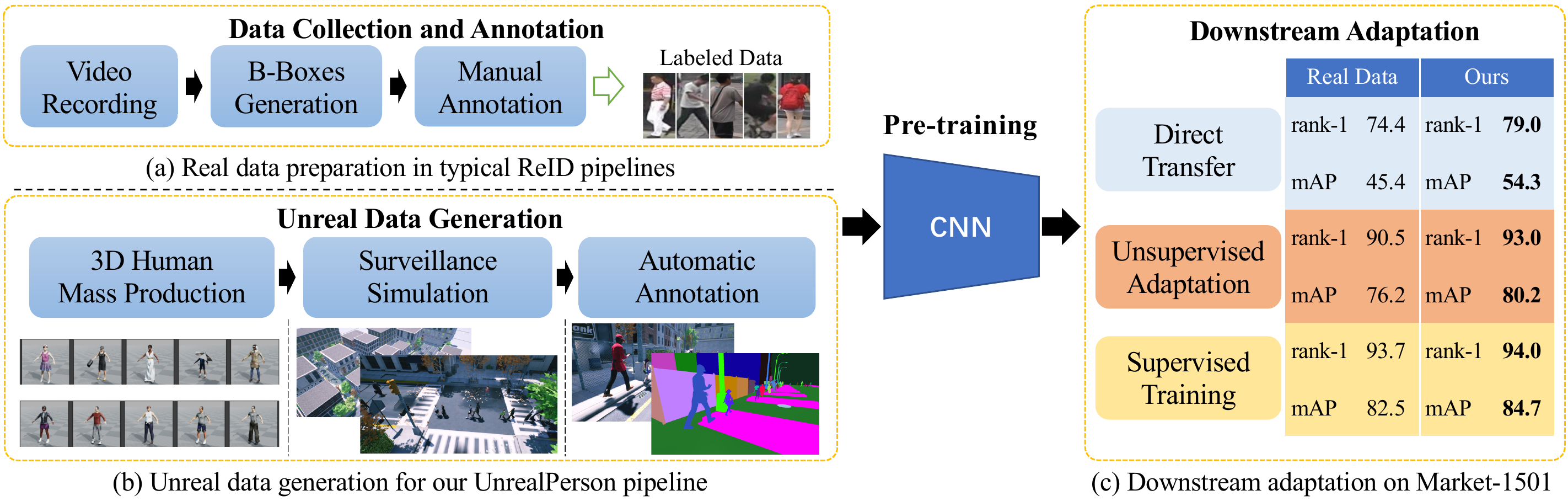}
    \caption{The UnrealPerson pipeline. The pre-training stage of our pipeline utilizes data synthesis, while conventional pipelines usually need annotated real-world data. We compare the best performance achieved by real data on the Market-1501 dataset. On all three tasks, our unreal data surpasses real data.}
    \label{fig:framework}
\end{figure*}

\subsection{Synthesized ReID Data}
Recently, some researchers adopt data synthesis techniques in ReID tasks.
SOMAset~\cite{somaset} is proposed to assist deep CNNs training. It contains only $50$ persons.
SyRI~\cite{SyRI} has $100$ persons under $140$ different lighting conditions.
These two datasets are rather small, and the diversity of backgrounds and human appearance is very limited.
PersonX~\cite{PersonX} is a large-scale synthesized dataset, containing $1\rm{,}266$ persons of multiple viewpoints. This dataset aims to explore how viewpoints affect ReID systems.
PersonX adopts ready-made human models from a 3D human dataset; thus, the expansibility is poor.
RandPerson~\cite{RandPerson} is a recent dataset proposed to generalize current ReID models. It contains a maximum of $8\rm{,}000$ pedestrians of $19$ cameras. By combining RandPerson and other real-world data as the training set, the ReID models achieve better supervised learning results.
Although RandPerson is more diverse and flexible for its human generation pipeline, it is still weaker than real-world datasets in terms of generalization ability.
In the following sections, we will discuss how to improve the quality of synthesized data. Our proposed insights make the synthesized data surpass real-world datasets and achieve the best performance on multiple ReID tasks. 

\section{The UnrealPerson Pipeline}
\subsection{Problem: Person Re-identification}

Given an annotated image dataset $\mathcal{S}=\{(\boldsymbol{I}_{1}, \boldsymbol{y}_{1}), (\boldsymbol{I}_{2}, \boldsymbol{y}_{2}), ..., (\boldsymbol{I}_{N}, \boldsymbol{y}_{N})\}$, where each $\boldsymbol{I}_i$ represents the image, and $\boldsymbol{y}_i$ is the ground truth label of the identity, the goal of ReID is to learn a proper feature embedding function $\boldsymbol{f}(\theta;\boldsymbol{I}_i)$ that maps images into a feature space $\mathcal{X}=\{\boldsymbol{x}_i|\boldsymbol{x}_i=\boldsymbol{f}(\theta;\boldsymbol{I}_i),1\leq i \leq N\}$, where the distances of the same identity are smaller than those of different identities. A straight-forward way to achieve this is to minimize the identity prediction error on $\mathcal{S}$:
\begin{equation}
    \min ~\mathbb{E}_{(\boldsymbol{I}_{i}, \boldsymbol{y}_{i})\in \mathcal{S}}[\boldsymbol{y}_{i}-\boldsymbol{g}(\boldsymbol{f}(\boldsymbol{I}_{i})],
\end{equation}
where $\boldsymbol{g}$ is the classifier. In this formulation, the quality of the learnt mapping relies on the data distribution of $\mathcal{S}$. The data distribution can be disassembled into two parts, the identity appearance distribution and background distribution. Hence, we denote the data distribution of $\mathcal{S}$ as $\mathcal{D}^{\mathcal{S}}(
\mathrm{FG},\mathrm{BG})$, where FG and BG represent foreground and background, respectively.

In the above formulation, two drawbacks are revealed. \textbf{First}, a major difficulty lies in collecting and annotating the training dataset $\mathcal{S}$.
A large-scale dataset often takes much time and labor to construct.
For instance, MSMT17 is a large-scale ReID dataset, consisting of $4\rm{,}101$ identities captured from $15$ cameras.
Researchers collected 180 hours of high-resolution videos and three labelers annotated for two months.
Such costs are unbearable in application scenarios.
\textbf{Second}, the data distribution $\mathcal{D}^{\mathcal{S}}$ is easily interfered with by foreground and background changes.
Thus, ReID data collected in one scene often fails to transfer well to other scenes.
For example, a ReID model trained on Market-1501 reports $91.4\%$ rank-1 accuracy when evaluating on Market-1501 testing set, where cameras are the same as the training set, but only obtains $25.7\%$ rank-1 accuracy on MSMT17.
This dramatic accuracy drop implies the huge domain gap and also reveals the weakness of the current ReID pipeline.


\subsection{Towards a Generalized and Costless Pipeline}


In the current ReID pipeline that involves data annotation from real scenes (as shown in Fig.~\ref{fig:framework}(a)),
the two drawbacks mentioned above are inevitable.
Usually, the costs of annotating a lot of cross-camera pedestrians are unbearable in applications.
To say the least, even if we do not consider the annotation costs, there is still a dilemma in real data preparation.
On the one hand, to offer sufficient coverage of different scenarios, the dataset should contain a large amount of labeled data.
On the other hand, a large amount of data may scatter data distribution, leading to an unsatisfying performance in some corner scenarios.
We owe such a dilemma to the real data lacking flexibility and turn to generating unreal data for training stronger ReID systems.



Unlike real data, synthesized data enjoys the benefits of free annotation and flexible data distribution.
Based on the toolkit of synthesizing data, we can easily pre-train a ReID model on an arbitrary distribution and, if needed, fine-tune it to adjust various downstream tasks.
As shown in Fig.~\ref{fig:framework}, our proposed pipeline involves three components: unreal data generation, model pre-training, and downstream adaptation, of which the unreal data is the fundamental part of the whole pipeline.
\begin{table*}[!ht]
  \centering

  \resizebox{0.98\textwidth}{!}{
    \setlength{\tabcolsep}{0.12cm}
    \begin{tabular}{l|c|c|c|c|c|c|c|c|c}
    \thickhline
    Datasets & \multicolumn{1}{l|}{\#Identities} & \multicolumn{1}{l|}{\#Cameras} & \multicolumn{1}{l|}{\#BBoxes} & Clothing & Accessories & Hard samples & \tabincell{c}{Surveillance\\Simulation} & Scalable & \multicolumn{1}{l}{\tabincell{c}{Rank-1 on \\MSMT17 (\%)}} \\
    \hline
    \hline
    SOMAset~\cite{somaset} & 50    & -     & 100,000 & Real  & No    & No    & No    & No    &  3.1\\
    SyRI~\cite{SyRI}  & 100   & -     & 56,000 & Real  & No    & No    & No    & No    & 21.8  \\
    PersonX~\cite{PersonX} & 1,266 & 6     & 273,456 & Real  & No    & No    & No    & Yes    & 22.2  \\
    RandPerson~\cite{RandPerson} & 8,000 & 19    & 228,655 & Generated+Real & No    & No    & Yes   & Yes   & 20.0  \\
    \hline
    Our synthesized data & 3,000 & 34    & 120,000 & Real  & Various & Many  & Yes   & Yes   & \textbf{38.5} \\
    \thickhline
    \end{tabular}%

}
    \caption{Detailed comparisons of synthesized datasets. Note that SOMAset and SyRI do not have a camera network, so numbers of cameras are left blank. The rank-1 accuracy on MSMT17 is the direct transfer performance of ReID models trained on these synthesized datasets.}
    \label{tab:comparison}
\end{table*}%

Our UnrealPerson pipeline liberates ReID systems from the manual annotation.
In terms of time cost for data preparation, we are able to construct a labeled unreal dataset of $3\rm{,}000$ identities within 48 CPU-hours, in comparison to the real dataset MSMT17 of $4\rm{,}101$ identities, which took 180 person-days.
This unreal dataset also surpasses MSMT17 on several downstream adaptation tasks, as shown in Fig.~\ref{fig:framework}(c).
Moreover, our pipeline is flexible compared to the typical ReID pipeline because the data distribution of synthesized data is fully controllable.
The flexibility of data synthesis empowers our pipeline to transfer well to many corner scenarios, like night-time ReID, indoor ReID, and so on, where annotated real data is hard to collect.
With our pipeline, we can easily fit these scenarios using unreal data.

In this paper, we mainly focus on dataset preparation and show that with a better-annotated dataset, the demands for pre-training methods and fine-tuning algorithms will be lowered greatly.
The details of our synthesized data will be discussed in the next subsection.

\subsection{Data Synthesis: The Devil Lies in Details}
We use synthesized data for our new ReID pipeline.
A data synthesis system is developed to generate costless and flexible ReID image data.
There are four major differences between our data synthesis system and others.
\textbf{(i) More realistic.} In foregrounds, we use real clothing images on generated 3D humans that comply with biological structures; in backgrounds, we mimic real surveillance systems in high-quality virtual environments.
Compared to PersonX or RandPerson that use low-poly assets in their generation systems, our assets are more realistic.
\textbf{(ii) More details.} We add more details to 3D human models. A total of $248$ types of clothes are used in our generated 3D humans. In addition, our 3D humans randomly carry accessories, including masks, glasses, hats, earphones, scarves, bags, and backpacks. These things are commonly seen in real scenes but hardly used in previous synthesis systems.
\textbf{(iii) Hard samples.} Apart from increasing diversity, we also consider the difficulty of the training set. Adding more details inevitably makes humans easy to recognize. Therefore, we deliberately add pedestrians with similar appearances in our synthesized data as hard samples. Persons that look quite similar but differs in small discriminative regions play an important role in guiding ReID models to focus on local areas.
\textbf{(iv) Scalability.} Different from SyRI and PersonX, our synthesis system supports arbitrary numbers of pedestrians and cameras because we have a 3D human production tool and a universal program that fits almost all virtual scenes in Unreal Engine 4.
The summarized comparisons are shown in Tab.~\ref{tab:comparison}.
We will validate the advantages mentioned above in Sec.~\ref{sec:dt}.
In the rest of this subsection, we briefly introduce the steps to generate the synthesized ReID data.

\noindent\textbf{3D Human Mass Production.}
The 3D human models are produced in MakeHuman, an open-source program that generates realistic human models.
We overwrite a community plugin, \textit{massproduce}~\cite{massproduce}, to generate a large number of models in one click.
To increase the diversity of human models, we use real-world clothing images of two datasets, Clothing-co-parsing~\cite{ccp} and DeepFashion~\cite{deepfashion}, as the clothing texture images for the generated humans.

\noindent \textbf{Surveillance Simulation.}
We simulate real surveillance scenes in Unreal Engine 4 (UE4), a mature platform for high-quality video games and VR applications.
The resources for UE4 are sufficient to generate various ReID datasets.
We choose 4 scenes from UE4 marketplace, namely Scene 1, ..., Scene 4, among which three are outdoor city environments, and one is an indoor scene.
For our virtual humans, we provide 4 walking animations and 2 idling animations.
They are given pre-defined paths to walk along in the unreal scenes.
Humans may occlude each other, just like in real-world scenes.
For cameras, we set virtual cameras in the unreal environments of different viewpoints and different distances to humans.
Multiple views of pedestrians can be obtained.
We also make skylight change during data collection.

\noindent\textbf{Data Annotation.}
Data annotation is conducted automatically by our annotating scripts.
We adopt UnrealCV~\cite{qiu2016unrealcv,qiu2017unrealcv} to collect pixel-level instance segmentation annotations for every image the virtual cameras capture.
Then, we crop every pedestrian in the images after filtering out small bounding boxes on the edge and discarding seriously occluded pedestrians.
To simulate detection bounding boxes or manually cropped boxes, the bounding boxes are randomly enlarged by a factor of $0.1$.


\begin{table*}[ht]
  \centering
  \resizebox{0.98\textwidth}{!}{
   \setlength{\tabcolsep}{0.16cm}
    \begin{tabular}{c|c|ccc|c|c|cc|cc|cc}
    \thickhline
    \multirow[c]{2}[0]{*}{\#IDs} & \multirow[c]{2}[0]{*}{\#Cameras} & \multicolumn{3}{c|}{Clothing Textures} & \multirow[c]{2}[0]{*}{Accessories} & \multirow[c]{2}[0]{*}{Hard Samples} & \multicolumn{2}{c|}{Market} & \multicolumn{2}{c|}{Duke} & \multicolumn{2}{c}{ MSMT17} \\
\cline{3-5}\cline{8-13}          &       & \multicolumn{1}{c}{Random} & \multicolumn{1}{c}{Generated} & Real  &       &       & rank-1 & mAP   & rank-1 & mAP   & rank-1 & mAP \\
    \hline
    \hline
    \multirow{5}[1]{*}{800} & \multirow{5}[1]{*}{6} & \multicolumn{1}{c}{\checkmark } &       &       &       &       & 52.0  & 26.2  & 41.4  & 22.4  & 18.5  & 6.1  \\
          &       &       & \multicolumn{1}{c}{\checkmark } &       &       &       & 61.0  & 34.4  & 49.8  & 29.8  & 19.6  & 6.6  \\
          &       &       &       & \checkmark  &       &       & 64.5  & 37.9  & 54.3  & 33.8  & 20.7  & 6.9  \\
          &       &       &       & \checkmark  & \checkmark  &       & 65.3  & 38.1  & 57.0  & 36.3  & 21.6  & 7.4  \\
          &       &       &       & \checkmark  & \checkmark  & \checkmark & 65.2  & 38.8  & 56.7  & 36.6  & 21.9  & 7.6  \\
          \hline
    \multirow[c]{4}[0]{*}{800} & 16    &       &       & \checkmark  & \checkmark  & \checkmark & 69.9  & 42.5  & 61.0  & 38.4  & 26.3  & 9.0  \\
          & 22    &       &       & \checkmark  & \checkmark  & \checkmark & 71.1  & 43.9  & 61.9  & 41.1  & 30.3  & 10.7  \\
          & 28    &       &       & \checkmark  & \checkmark  & \checkmark & 73.7  & 46.5  & 62.9  & 41.6  & 33.4  & 12.6  \\
          & 34    &       &       & \checkmark  & \checkmark  & \checkmark & 74.9  & 48.2  & 64.9  & 42.9  & 35.4  & 13.3  \\
    \hline
    1,500 & \multirow[c]{3}[0]{*}{34} &       &       & \checkmark  & \checkmark  & \checkmark & 75.7  & 50.7  & 67.3  & 46.6  & 36.3  & 13.9  \\
    2,000 &       &       &       & \checkmark  & \checkmark  & \checkmark & 76.8  & 52.0  & 69.0  & 48.0  & 37.9  & 14.7  \\
    3,000 &       &       &       & \checkmark  & \checkmark  & \checkmark & \textbf{79.0} & \textbf{54.3} & \textbf{69.7} & \textbf{49.4} & \textbf{38.5} & \textbf{15.3} \\
    \thickhline
    \end{tabular}%
    }
    \caption{Direct transfer performance of Unreal to three real datasets. We control several parameters in our synthesized data,~\emph{i.e.}, clothing textures, accessories, hard samples and numbers of identities and cameras, towards better performance. }
  \label{tab:dtall}%
\end{table*}%

\noindent\textbf{Summary.} Our synthesized dataset sets up a better platform for downstream tasks, and even the pre-trained model itself achieves good results on many datasets.
The UnrealPerson pipeline lowers the needs for annotated real data and boosts transferring performance across different domains, which will be presented in the next section.
More technical details about data synthesis and visualizations of our synthesized data are shown in supplementary materials.

\section{Experiments}
In this section, after introducing the implementation details, we validate our UnrealPerson pipeline from three aspects. First, we evaluate the quality of our synthesized data via direct transfer. Next, we conduct experiments on domain adaptation. Last, we show the convenience of data distribution adjustment in our pipeline by evaluating on several corner scenarios.

\subsection{Implementation Details}
We adopt ResNet-50~\cite{ResNet},  which is pre-trained on ImageNet~\cite{imagenet_cvpr09}, as our backbone for all experiments.
We also replace all batch normalization layers with camera-based batch normalization (CBN)~\cite{CBN} layers in the network.
An extra CBN layer is added after global average pooling on the last residual block of ResNet-50, followed by a linear layer as the classifier.
To train with labeled data, the softmax cross-entropy loss is used.
For unlabeled data, we use the joint visual and temporal consistency (JVTC)~\cite{JVTC} framework for unsupervised domain adaptation.
The images are resized to $256\times128$. For direct transfer experiments, we set batch size as $64$; for unsupervised domain adaptation experiments, we sample $128$ images from the source domain and $128$ images from the target domain to form a mini-batch. In each mini-batch, for annotated data, we adopt a balanced sampling strategy proposed in ~\cite{SCT}.
For unlabeled data, we randomly sample images. We adopt the SGD optimizer for training, with a momentum of $0.9$ and weight decay of $5\times10^{-4}$. The initial learning rate is $0.01$ and decays to $0.001$ after $40$ epochs. The training stops at the $60th$ epoch.
Besides, as synthesized datasets contain more images, we observe convergence with fewer iterations. Therefore, for synthesized datasets, in direct transfer experiments, we decay the learning rate at the $10th$ epoch and stop training at the $15th$ epoch; in unsupervised domain adaptation experiments, we only sample $300$ mini-batches from the source domain in each epoch.

\subsection{Direct Transfer Evaluation} \label{sec:dt}
The synthesized ReID data is the fundamental part of the UnrealPerson pipeline.
Here, we adopt direct transfer performance on real datasets as the indicator to show the quality of synthesized data because direct transfer is the basis for all other tasks.
Three real datasets, Market-1501~\cite{MARKET}, DukeMTMC-reID~\cite{DUKEMTMC}, and MSMT17~\cite{PTGAN}, are used as the testing sets. For short, we refer to Market-1501 and DukeMTMC-reID as Market and Duke, respectively.
For our synthesized data, we refer to as Unreal and add suffixes to Unreal to distinguish different variants.
We sample $40$ images for each pedestrian to form our Unreal dataset.
We first explore how to synthesize high-quality virtual humans and then scale up the number of cameras and identities. A summarized report is presented in Tab.~\ref{tab:dtall}.
Next, we describe every point in Tab.~\ref{tab:dtall} separately.

\begin{figure}[!t]
\centering
\includegraphics[width=0.44\textwidth]{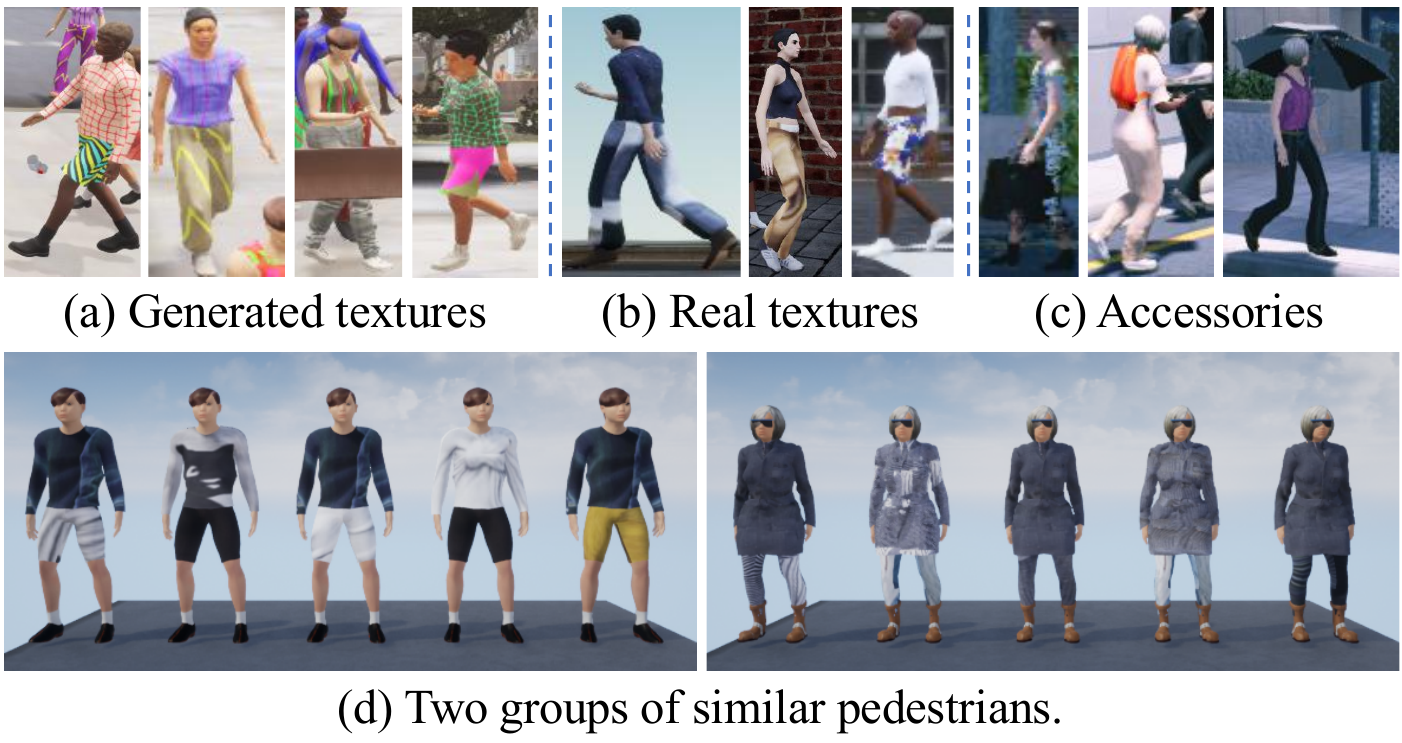}
\caption{Visualizations of our synthesized data. (a) Regular generated clothing textures. (b) Real textures. (c) 3D humans with accessories (handbag, backpack, and umbrella). (d) Pedestrians with similar appearance.}
\label{fig:unrealdata}
\end{figure}

\noindent \textbf{Clothing Textures.} The appearance of clothes is a main part of the foreground and provides much discriminative information.
In synthesized data, the clothing templates are fixed, limiting the richness of clothing appearance.
Therefore, we randomly replace the clothing textures to address this problem.
We compare three different ways of enriching clothing textures. Random images are from universal image datasets, like ImageNet~\cite{imagenet_cvpr09} or COCO~\cite{coco}. Generated color textures (see Fig.~\ref{fig:unrealdata}(a)) are proposed in RandPerson~\cite{RandPerson}. A few pre-defined patterns are applied to generated color palettes, producing lots of generated regular textures.
Real textures are cropped clothing image patches from clothing segmentation datasets.
From the top 3 lines in Tab.~\ref{tab:dtall}, we can conclude that the textures of real clothes are more suitable for enriching 3D human models.
Random images or generated patterns may vary from real clothes largely.
Although they also increase the diversity of human appearance, the reality of unreal data is decayed in the meantime.

\noindent\textbf{Accessories.}
The accessories of pedestrians are important clues to recognize their identities.
Adding accessories to 3D humans makes the synthesized data more realistic.
As shown in Fig.~\ref{fig:unrealdata}(c), our data synthesis system supports various accessories on human models.
From Tab.~\ref{tab:dtall}, we can see that the synthesized datasets with accessories achieve better performance on all three testing sets. For example, on Duke, the unreal data with accessories improves the rank-1 accuracy from $54.3\%$ to $57.0\%$.

\begin{figure*}[!t]
\centering
\subfigure[Evaluation on Market]{
\includegraphics[height=0.2\textwidth]{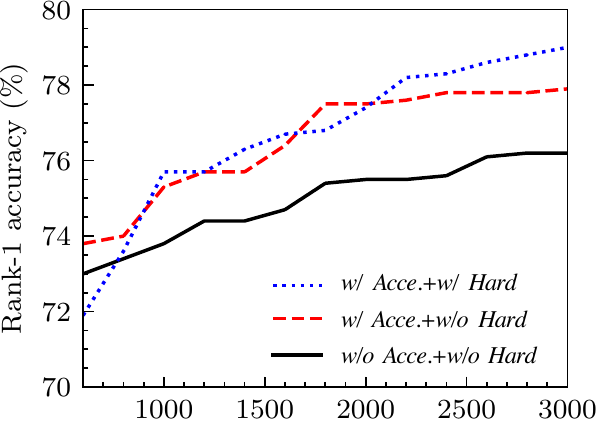}
}
\subfigure[Evaluation on Duke]{
\includegraphics[height=0.2\textwidth]{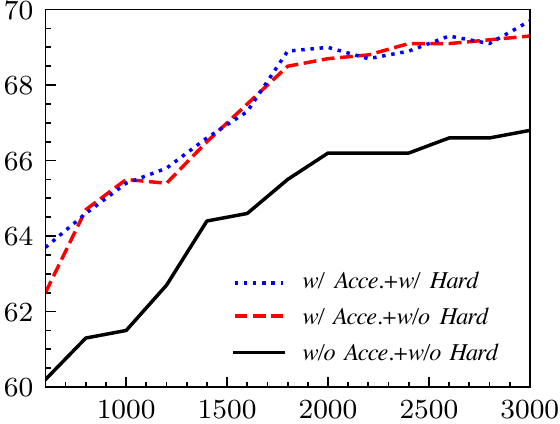}
}
\subfigure[Evaluation on MSMT17]{
\includegraphics[height=0.2\textwidth]{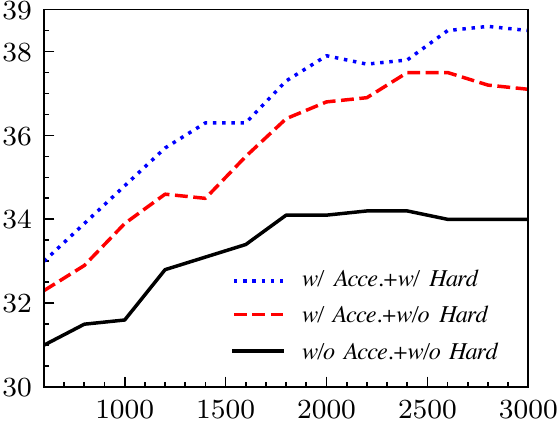}
}
\caption{Results of direct transfer evaluation on Market, Duke and MSMT17, with the number of pedestrians in synthesized datasets increasing from $600$ to $3\rm{,}000$. The number of cameras in all the experiments shown in this figure is $34$.}
\label{fig:personnum}
\end{figure*}

\noindent\textbf{Hard Samples.}
In real-world scenarios, some persons may look quite similar.
They are \textit{hard samples} for ReID algorithms and are important for efficient learning.
As shown in Fig.~\ref{fig:unrealdata}(d), we also synthesize hard samples for our unreal data.
Specifically, we generate human models by group.
The five human models in one group share similar appearances, but everyone is a little different from the other four.
Note that hard samples are better provided with a rather larger number of identities and cameras.
In a small dataset, hard samples may heavily interfere learning from normal cases.
When abundant normal samples are provided, they take a positive effect instead.
As shown in Tab.~\ref{tab:dtall}, when we have only $800$ identities in $6$ cameras, the hard samples slightly improve mAP on the three testing sets.
From Fig.~\ref{fig:personnum}, when we use more cameras and more identities, adding these similar pedestrians significantly improves the rank-1 accuracy on MSMT17.

\begin{figure}[!t]
\centering
\includegraphics[width=0.43\textwidth]{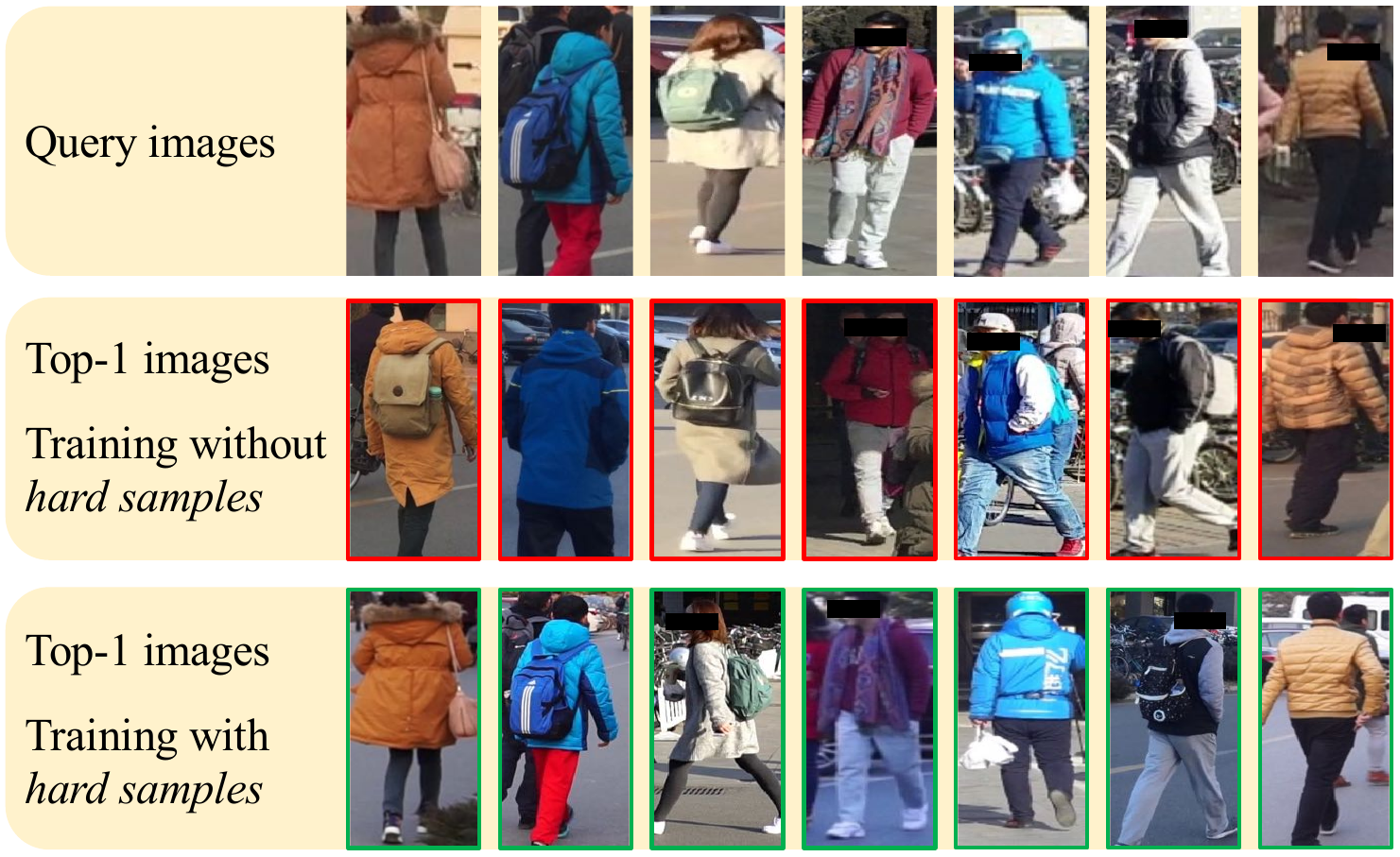}
\caption{Top-1 results on MSMT17. The ReID models are trained using our unreal data with or without hard samples,~\emph{i.e.}, similar pedestrians. Images in the same column are interrelated. Red: false matches; Green: correct matches.}
\label{fig:simi}
\end{figure}

We also show some query results on MSMT17 in Fig.~\ref{fig:simi}.
The ReID model trained with synthesized data without hard samples tends to ignore the obvious differences in backpacks and handbags.
After adding similar pedestrians as hard samples, we empower ReID methods to focus on local regions, achieving better accuracy in real-world scenes.

\noindent\textbf{Number of cameras.}
We prepare $4$ scenes for the 3D human models to walk around, where $34$ virtual cameras are deployed to capture images.
We construct several synthesized datasets containing $800$ humans with the number of cameras increasing from $6$ to $34$ and involving scenes from $1$ to $4$.
The direct transfer evaluation results are shown in the middle part of Tab.~\ref{tab:dtall}.
Virtual environments differ from each other in many aspects, such as illumination, styles of buildings and roads, and crowdedness.
By recognizing persons across virtual environments, ReID models are more robust to person-unrelated variations of different scenes.

\noindent\textbf{Number of pedestrians.}
Apart from the results shown in the bottom part of Tab.~\ref{tab:dtall}, we also conduct fine-grained experiments to explore how many pedestrians are suitable for training ReID models.
The accessories and hard samples are also validated in this series of experiments.
From Fig.~\ref{fig:personnum}, we can see that generally, more pedestrians lead to better performance. The results also demonstrate the effectiveness of our proposed additional components for foreground synthesis,~\emph{i.e.}, accessories and hard samples.
Note that we achieve the best direct transfer performance using $3\rm{,}000$ identities, much fewer than RandPerson.
We also observe that, when adding more pedestrians (larger than $3\rm{,}000$), the performance is hardly promoted. This issue will be discussed in Sec.~\ref{sec:op}.



\begin{table}[t]
  \centering

  \resizebox{0.47\textwidth}{!}{
    \begin{tabular}{l|cc|cc|cc}
    \thickhline
    \multirow[c]{2}[0]{*}{Training set} & \multicolumn{2}{c|}{Market} & \multicolumn{2}{c|}{Duke} & \multicolumn{2}{c}{ MSMT} \\
\cline{2-7}          & rank-1 & mAP   & rank-1 & mAP   & rank-1 & mAP \\
    \hline
    \hline
    Market & \textit{91.4}  & \textit{76.8}  & 56.7  & 36.5  & 25.7  & 9.6  \\
    Duke  & 72.4  & 41.9  & \textit{82.1}  & \textit{67.5}  & 35.8  & 13.1  \\
    MSMT17  & 74.4  & 45.4  & 67.1  & 46.8  & \textit{72.5}  & \textit{42.4}  \\
    \hline
    SyRI  & 48.5  & 22.6  & 38.9  & 18.2  & 21.8  & 5.7  \\
    PersonX & 58.7  & 32.7  & 49.4  & 28.9  & 22.2  & 7.9  \\
    RandPerson & 64.7  & 39.3  & 59.4  & 38.4  & 20.0  & 6.8  \\
    \hline
    Unreal & \textbf{79.0} & \textbf{54.3} & \textbf{69.7} & \textbf{49.4} & \textbf{38.5} & \textbf{15.3} \\
    \thickhline
    \end{tabular}%
}

      \caption{Direct transfer performance of real datasets and synthesized datasets. The fully-supervised learning results are in \textit{italics}.}
  \label{tab:direct}%
\end{table}%

\noindent\textbf{Summary.} In the above analysis, we explore the key factors of how to improve synthesized ReID datasets. In foreground synthesis, we generate 3D human models with accessories to introduce more diversity and produce similar humans as hard samples. In background synthesis, we validate the importance of cross-scene pedestrians.
Based on these conclusions, we take a representative synthesized dataset that exploits all the advantages we find. This dataset contains $120\rm{,}000$ images of $3\rm{,}000$ pedestrians collected from $34$ cameras deployed in $4$ different virtual scenes.
For convenience, we refer to this dataset as Unreal.
The direct transfer performance of Unreal and other datasets is compared in Tab.~\ref{tab:direct}.


\subsection{Domain Adaptation}

\noindent\textbf{Supervised Fine-tuning.}
When abundant labeled data in the target domain is accessible, domain adaptation can be simply done by supervised fine-tuning on the pre-trained model.
From Tab.~\ref{tab:pretrain}, we see that the results of fully-supervised learning are promoted by using Unreal as the pre-training data. For example, when training and testing on Market with ImageNet pre-trained model, the rank-1 accuracy is $91.4\%$, while our Unreal pre-trained model reaches $94.0\%$.
Our Unreal dataset also surpasses other real-world ReID datasets, showing its universality and transferability.

\begin{table}[t]
  \centering
    \resizebox{0.47\textwidth}{!}{
    \setlength{\tabcolsep}{0.12cm}

    \begin{tabular}{l|c|cc|cc|cc}
    \thickhline
    \multirow[c]{2}[0]{*}{\tabincell{l}{Pre-training \\Dataset}} & \multicolumn{1}{l|}{\multirow[c]{2}[0]{*}{\tabincell{l}{Fine-tuning\\Dataset}}} & \multicolumn{2}{c|}{Market} & \multicolumn{2}{c|}{Duke} & \multicolumn{2}{c}{ MSMT} \\
\cline{3-8}          &       & rank-1 & mAP   & rank-1 & mAP   & rank-1 & mAP \\
    \hline
    \hline
    ImageNet & \multirow[c]{3}[0]{*}{Market} & \textit{91.4} & \textit{76.8} & 56.7  & 36.5  & 25.7  & 9.6  \\
    MSMT17  &       & \textit{93.7} & \textit{82.5} & 68.5  & 50.1  & 47.3  & 21.4  \\
    Unreal &       & \textit{\textbf{94.0}} & \textit{\textbf{84.7}} & 72.8  & 54.7  & 35.6  & 14.8  \\
    \hline
    ImageNet & \multirow[c]{3}[0]{*}{Duke} & 72.4  & 41.9  & \textit{82.1} & \textit{67.5} & 35.8  & 13.1  \\
    MSMT17  &       & 76.5  & 47.6  & \textit{85.7} & \textit{72.1} & 48.5  & 20.6  \\
    Unreal &       & 82.5  & 57.9  & \textit{\textbf{86.8}} & \textit{\textbf{74.2}} & 40.5  & 16.3  \\
    \hline
    ImageNet & \multirow[c]{3}[0]{*}{MSMT17} & 74.4  & 45.4  & 67.1  & 46.8  & \textit{72.5} & \textit{42.4} \\
    Duke+Market &       & 80.0  & 53.6  & 70.5  & 52.6  & \textit{73.7} & \textit{44.7} \\
    Unreal &       & 80.1  & 53.8  & 71.5  & 52.8  & \textit{\textbf{74.5}} & \textit{\textbf{46.0}} \\
    \thickhline
    \end{tabular}%

}

      \caption{Results of supervised fine-tuning on the pre-trained models. The fully-supervised learning results are in \textit{italics}. Direct transfer performance is also shown in this table conveniently.}
  \label{tab:pretrain}%
\end{table}%


\begin{table}[t]
  \centering

\resizebox{0.47\textwidth}{!}{
    \setlength{\tabcolsep}{0.12cm}

    \begin{tabular}{l|c|cc|cc|cc}
    \thickhline
    \multirow[c]{2}[0]{*}{Source Domain} & \multirow[c]{2}[0]{*}{Methods} & \multicolumn{2}{c|}{Market} & \multicolumn{2}{c|}{Duke} & \multicolumn{2}{c}{ MSMT17} \\
\cline{3-8}          &       & rank-1 & mAP   & rank-1 & mAP   & rank-1 & mAP \\
    \hline
    \hline
    Market & \multirow[c]{4}[0]{*}{JVTC} & -     & -     & 76.5  & 59.6  & 46.1  & 20.4  \\
    Duke  &       & 89.0  & 73.1  & -     & -     & 52.5  & 23.5  \\
    MSMT17 &       & 89.9  & 74.5  & 79.0  & 63.5  & -     & - \\
\cline{1-1}\cline{3-8}    Unreal &       & \textbf{90.8} & \textbf{78.3} & \textbf{81.2} & \textbf{66.1} & \textbf{53.7} & \textbf{25.0} \\
    \hline
    \hline
    Market & \multirow[c]{4}[0]{*}{JVTC+} & -     & -     & 84.6  & 68.8  & 63.7  & 30.1  \\
   Duke  &       & 89.3  & 74.6  & -     & -     & 66.8  & 32.5  \\
   MSMT17 &       & 90.5  & 76.2  & 85.2  & 72.1  & -     & - \\
\cline{1-1}\cline{3-8}    Unreal &       & \textbf{93.0} & \textbf{80.2} & \textbf{88.3} & \textbf{75.2} & \textbf{68.2} & \textbf{34.8} \\
    \thickhline
    \end{tabular}%

}
    \caption{Unsupervised domain adaptation performance on three benchmark datasets.}
  \label{tab:uda}%
\end{table}%

\noindent\textbf{Unsupervised Domain Adaptation.}
Unsupervised domain adaptation (UDA) is a popular direction to leverage unlabeled data from the target domain. The training sets include labeled data of the source domain and unlabeled data from the target domain.
In UDA experiments, we implement JVTC~\cite{JVTC} as the off-the-shelf algorithm in the network with CBN~\cite{CBN} layers. JVTC+ denotes using joint similarity of visual and spatial-temporal features in the testing stage.
The results are shown in Tab.~\ref{tab:uda}.
With the assistance of our unreal data, the UDA performance is largely boosted.
On Duke, JVTC+ further promotes the rank-1 accuracy to $88.3\%$, not only setting up a state-of-the-art record but also surpassing the fully-supervised learning results shown in Tab.~\ref{tab:pretrain}.
Note that these results are obtained without any manual annotation.
It demonstrates the application values of our UnrealPerson pipeline.

\begin{figure*}
    \centering
    \includegraphics[width=0.96\textwidth]{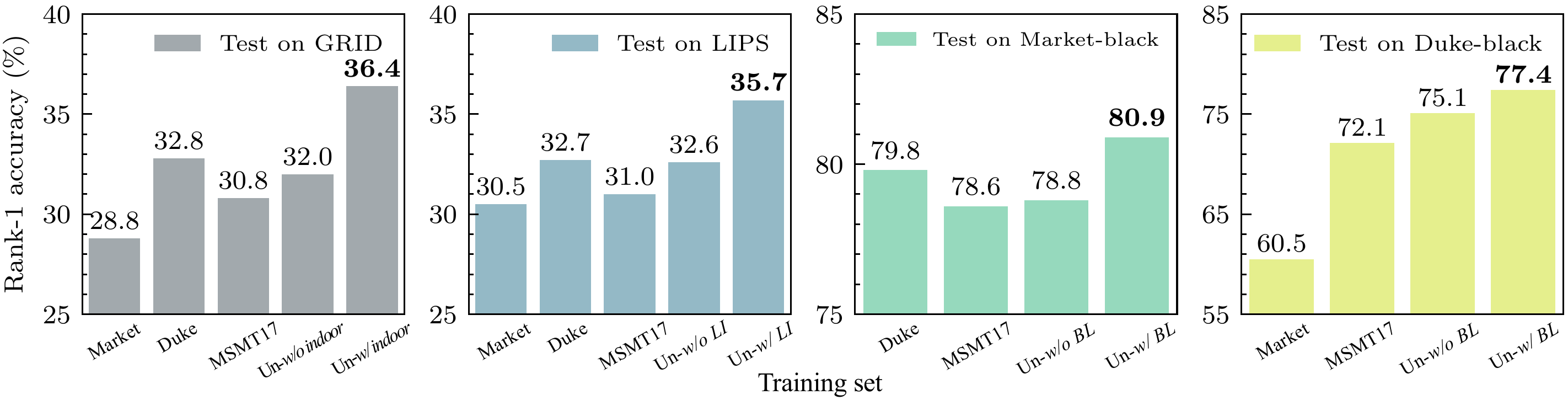}
    \caption{Direct transfer performance on various corner scenarios. \textit{Un} denotes the \textit{unreal} data are used for training.}
    \label{fig:corner}
\end{figure*}

\subsection{Task-specific Adaptation for Corner Scenarios}
Our UnrealPerson pipeline enjoys the benefits of flexible data synthesis.
The distribution of synthesized data can be adaptively adjusted by modifying parameters of synthesis, and the process can be done easily in our UnrealPerson pipeline.
This advantage makes our pipeline more suitable for corner scenarios of ReID, where labeled data is hard to obtain.
We present three examples,~\emph{i.e.}, indoor ReID, low illumination ReID and black ReID.
Previously little attention has been paid to these scenarios for the lack of abundant training data.
For evaluation on indoor ReID and low illumination ReID tasks, we use GRID~\cite{GRID} and LIPS~\cite{LIPS} dataset, respectively.
GRID is a dataset collected in an underground station, where the cameras are located at high angles of view.
LIPS is constructed with two night cameras, and thus the illumination is extremely low.
We take 250 persons in GRID and 50 persons in LIPS for testing.
Black ReID problem was first defined in~\cite{xu2020black}. In some scenes, most pedestrians wear similar clothes,\emph{e.g.}, many wear dark clothes in winter.
We prepare two datasets, Market-black and Duke-black, according to the annotations provided by the Black-reID dataset~\cite{xu2020black}. All persons in Duke-black and Market-black wear black clothes. Duke-black contains 1,216 images of 145 persons, among which 436 images are used as queries and the rest as gallery images. Market-black contains 41 persons. It has 174 images as queries and 251 images as the gallery.

In Fig.~\ref{fig:corner}, we show the direct transfer performance on these datasets. The compared training sets include Market, Duke, and MSMT17.
For the indoor dataset GRID, we use the Unreal dataset with 6 extra indoor cameras, achieving 36.4\% rank-1 accuracy. On the low illumination dataset, we adjust our virtual scenes to night time, and train the network with low illumination (LI) unreal data, surpassing real data by 3.0\%.
For black ReID, we apply dark clothing textures to 3D humans to construct our Unreal-\textit{w/ BL}. This dataset gets significant promotions on the two black datasets.

\section{Open Problems} \label{sec:op}

Our research leaves a few open problems for the community. We summarize them as follows, and hence suggest the community pays more attention to this new pipeline that is potentially the future trend of ReID.
\begin{itemize}
    \setlength{\itemsep}{0.02cm}
    \item The \textbf{quality} of synthesized data. It is easy to recognize the synthesized images from the real images, because the realness of 3D human models, the richness of facial expressions and contexts, the illumination changes, \textit{etc.}, are still far from perfect. We guess that there is a saturation point for synthesized data (beyond it, continuing improving reality brings marginal gains), but we are not sure when it will be reached and whether the domain transfer methods can relieve the burden of image synthesis. Within a short period, we believe that continuing mimicking the real-world data property can bring us non-trivial benefits.
    \setlength{\itemsep}{0.02cm}
    \item The \textbf{quantity} of synthesized identities. Currently, our algorithm reaches a plateau in direct transfer at $3\rm{,}000$ pedestrians. Surprisingly, this number is even smaller than MSMT17, the real-world ReID dataset. There is no doubt about the potential of introducing more data, but there are problems to solve, including the quality issue (described above) and the data distribution issue (\textit{e.g.}, the number of identities to be placed in one scenario, the function that samples the details for each identity, \textit{etc.}). Clues may be found by analyzing some meta-information (\textit{e.g.}, the distribution of identity similarity) and compare it to real-world datasets.
    \setlength{\itemsep}{0.02cm}
    \item The \textbf{efficiency} of learning from infinite data. Prior works~\cite{sener2018active,chen2018sampleahead} have shown that active learning or hard example mining are potentially more efficient strategies when there is an infinite amount of data. We hope to validate these techniques in our pipeline and thus decrease the complexity, in particular when the data quantity becomes much larger.
    \setlength{\itemsep}{0.02cm}
    \item The \textbf{relationship} with other problems. The flexibility of data synthesis allows us to augment the scope of ReID, or investigate the relationship between ReID and other vision problems. To name a few, (i) one can generalize ReID from image-based to video-based, where our pipeline enjoys a larger advantage over the public benchmarks; (ii) one can generate high-quality segmentation mask for other objects in the virtual scenarios, allowing the researchers to consider the contexts for more accurate ReID; (iii) one can also study the self-supervised learning methods~\cite{moco,chen2020simple,chen2020big}, which often require more data and are believed stronger in domain transfer.
\end{itemize}

\section{Conclusions}

This paper presents \textbf{UnrealPerson}, a novel pipeline for person re-identification (ReID). It aims to relieve the burden of costly data annotation and alleviate the difficulty of domain transfer. Synthesized data plays an important role in this research. We reveal that there is still much room of improvement by synthesizing data from more virtual scenarios/cameras as with richer details. With our pre-trained ReID model, the direct transfer accuracy to MSMT17, the largest publicly available dataset, is almost doubled compared to the previous best pipeline that uses synthesized training data. UnrealPerson enjoys stronger transferability to real-world ReID datasets because (i) the pre-trained model is specialized in and better at processing ReID data, and (ii) the synthesized environment can be flexibly adjusted to the corner scenarios in which collecting real-world data is difficult.

{\small
\bibliographystyle{ieee_fullname}
\bibliography{egbib}
}

\end{document}